\documentclass{article}

\usepackage{microtype}
\usepackage{graphicx}
\usepackage{subfigure}
\usepackage{booktabs} % for professional tables

\usepackage{hyperref}

\usepackage{enumitem}
\setlist{nosep}
\usepackage{array}
\usepackage[table]{xcolor}
\usepackage{soul}
\definecolor{lightred}{RGB}{255, 204, 204}
\definecolor{lightgreen}{RGB}{204, 255, 204}
\definecolor{yellowgreen}{RGB}{240, 250, 140}
\definecolor{lightblue}{RGB}{204, 229, 255}

\usepackage[accepted]{icml2025}

\usepackage{amsmath}
\usepackage{amssymb}
\usepackage{mathtools}
\usepackage{amsthm}

\usepackage[capitalize,noabbrev]{cleveref}

\definecolor{tcboxcolor}{RGB}{242,245,237}
\definecolor{framecolor}{RGB}{179,191,166}
\usepackage{tcolorbox}

\theoremstyle{plain}
\newtheorem{theorem}{Theorem}[section]

\theoremstyle{definition}
\newtheorem{definition}[theorem]{Definition}

\theoremstyle{remark}

\usepackage[textsize=tiny]{todonotes}

\icmltitlerunning{Agentic Systems Theory}

\begin{document}

\twocolumn[
\icmltitle{Agentic AI Needs a Systems Theory}




\begin{icmlauthorlist}
\icmlauthor{Erik Miehling}{}
\icmlauthor{Karthikeyan Natesan Ramamurthy}{}
\icmlauthor{Kush R. Varshney}{}
\icmlauthor{Matthew Riemer}{}
\icmlauthor{Djallel Bouneffouf}{}
\icmlauthor{John T. Richards}{}
\icmlauthor{Amit Dhurandhar}{}
\icmlauthor{Elizabeth M. Daly}{}
\icmlauthor{Michael Hind}{}
\icmlauthor{Prasanna Sattigeri}{}
\icmlauthor{Dennis Wei}{}
\icmlauthor{Ambrish Rawat}{}
\icmlauthor{Jasmina Gajcin}{}
\icmlauthor{Werner Geyer}{}\\
\vspace{0.5em}{\centering IBM Research}
\end{icmlauthorlist}

%\icmlaffiliation{ibm}{IBM Research}

\icmlcorrespondingauthor{Erik Miehling}{erik.miehling@ibm.com}
\icmlcorrespondingauthor{Karthikeyan Natesan Ramamurthy}{knatesa@us.ibm.com}

\icmlkeywords{}

\vskip 0.3in
]



\printAffiliationsAndNotice{}  % leave blank if no need to mention equal contribution

\begin{abstract}
The endowment of AI with reasoning capabilities and some degree of agency is widely viewed as a path toward more capable and generalizable systems. Our position is that the current development of agentic AI requires a more holistic, systems-theoretic perspective in order to fully understand their capabilities and mitigate any emergent risks. The primary motivation for our position is that AI development is currently overly focused on individual model capabilities, often ignoring broader emergent behavior, leading to a significant underestimation in the true capabilities and associated risks of agentic AI. We describe some fundamental mechanisms by which advanced capabilities can emerge from (comparably simpler) agents simply due to their interaction with the environment and other agents. Informed by an extensive amount of existing literature from various fields, we outline mechanisms for enhanced agent cognition, emergent causal reasoning ability, and metacognitive awareness. We conclude by presenting some key open challenges and guidance for the development of agentic AI. We emphasize that a systems-level perspective is essential for better understanding, and purposefully shaping, agentic AI systems.
\end{abstract}

\section{Introduction}
\label{sec:intro}

% outline various efforts on agentic AI and their promises

Agentic AI systems, which aim to solve long-horizon tasks through sophisticated reasoning with minimal human supervision, have become a central focus of current AI development. Recent research advances have accelerated progress in this direction \cite{li2024survey,acharya2025agentic}, with major labs pushing to develop increasingly autonomous agents \cite{anthropic_computer_2024,anthropic_claude_code_2025,openai2025operator,deepmind2025mariner}. The promises of agentic AI are significant, from assisting business operations \cite{chawla2024agentic} to automating clinical workflows \cite{qiu2024llm} to advancing scientific research \cite{lu2024ai}.

% list challenges, foreshadow the primary questions we will aim to target in the position paper

Unsurprisingly, there are numerous challenges with building effective agentic AI. The problem solving abilities of current LLM-based agents are significantly limited, especially in longer horizon tasks, primarily due to their difficulty in interfacing with the environment (and humans), lack of commonsense, and even tendency toward self-deception \cite{xu2024theagentcompany}. Broadening scope to domains where an agent must communicate with humans, interact with other agents, and deal with the full complexities of operating in the wild (e.g., acting in non-stationary domains), %sensing and actuating as an autonomous physical agent), 
the task of building robust and safe AI agents becomes an even greater challenge. These agents face various obstacles including acting under fundamental uncertainty (and incompleteness) in their world models \cite{vafa2024evaluating}, fulfilling goals while maintaining task corrigibility/flexibility and appropriate bounds on agency \cite{chan2023harms}, interacting (both cooperatively and competitively) with other agents  \cite{tran2025multi}, and effectively communicating information to (and receiving feedback from) users \cite{bansal2024challenges}, all while being sure to operate within the rules, regulations, and ethical norms of human institutions \cite{rao2023ethical, shavit2023practices, kolt2024governing}.

%interacting with other agents

%emergent behavior at the system level

%system level considerations

%resolving dilemmas with complex ethical and moral reasoning \cite{rao2023ethical}.

\iffalse
- more broadly, when we allow AI to create sub-goals, how do we ensure that those subgoals align with the constraints imposed by humans % Hinton

How do we ensure that humans ``remain in the driver's seat'' when AI becomes sufficiently capable and intelligent?
\fi

% argue that the solutions are unstructured; claim that the field would benefit from a common theoretical language 

{This position paper argues that the development of agentic AI requires a holistic, systems-theoretic perspective to fully understand their capabilities and mitigate emergent risks.} % interrelated -- and often competing -- demands of capability, controllability, and institutional compliance.} 
Our position's primary motivation (and our main concern) is that AI development is currently overly focused on capabilities in isolation, often ignoring broader systemic considerations. We argue that being overly focused on model capabilities leads the community to underestimate both the true capabilities and the associated risks of agentic AI. This capabilities-centric approach has already produced some concerning emergent behaviors. Recent experiments show that Anthropic's Claude demonstrates deceptive behavior, termed \emph{alignment faking}, in which the model will exhibit a particular behavior during training or when monitored, only to revert to different, often disallowed behaviors once that oversight is absent \cite{greenblatt2024alignment}. Other research demonstrates that some modern models may make attempts to ``steal [their] own weights'' (in a process termed ``self-exfiltration'') when given the opportunity, intentionally ``sandbag'' performance when threatened with unlearning, or disable oversight mechanisms if the mechanism interferes with achieving a goal \cite{meinke2024frontier}. Early implementations of agentic AI, in the context of a simulated workplace, have demonstrated that agents may deceive themselves into (falsely) satisfying goals \cite{xu2024theagentcompany}, e.g., an agent who was unable to find a particular user ended up creating a ``shortcut solution by renaming another user to the name of the intended user.'' The above cases were all observed in highly controlled (simulated) environments; as models become more capable and further integrated into society, these behaviors will become much more complex and increasingly difficult to detect and control.

Systems theory \cite{wiener1948cybernetics,boulding1953organizational,ashby1956introduction,bertalanffy1968general,astrom2008feedback} --- the general study of how complex wholes emerge from the interactions of their constituent parts --- stresses how each component of a system must be understood both in terms of its individual definition and its contribution to the larger system's behavior. Systems theories exist in a variety of fields, from biology's understanding of cellular networks, to sociology's analysis of social and organizational structures, to engineering's development of control systems. Agentic systems, consisting of agents iteratively interacting with humans and other agents to achieve specified tasks, possess properties that are amenable to a systems-level analysis. At the most granular level, an agent contains an internal act-sense-adapt loop. This loop feeds, and is fed by, feedback loops at higher levels, namely at the agent-human interface, the agent-agent interface, and the agent-environment interface. These complex interactions can lead to fundamentally different behavior at the system level. In particular, as we will discuss, there are several viable mechanisms of emergence that can allow the system to exhibit advanced causal reasoning capabilities and metacognitive awareness, even though the internal processes of agents are much simpler. This allows the system as a whole to possess a type of (collective) agency. 

Our position aims to develop an \emph{agentic AI systems theory} to describe how agency at the system level can emerge from the interactions between much simpler agents (tool-use LLMs), humans, and the environment. The development of this theory naturally draws upon fields beyond the AI community, namely psychology, neuroscience, cognitive science, sociology, and biology. Our position does \emph{not} claim that a systems theory will yield immediate solutions for the current risks of agentic AI systems, but rather that we as a community should be more intentional about considering the emergent capabilities of agentic AI instead of focusing solely on model capabilities. Additionally, we do not necessarily advocate for the construction of increasingly agentic systems; our goal is to help the community better understand the emergent behavior of agentic AI so that we can 
design tools to facilitate more deliberate design of their capabilities. Our paper aims to take a meaningful step in this direction.

{\bf Related Work.} The body of work on agentic AI is rapidly growing; we focus on some of the most relevant work below.

\emph{Human-AI interaction.} The interactions between humans and AI can be incredibly complex. The work of \cite{mitelutposition} provides insight into this complexity by introducing the concept of ``agency loss": the phenomenon that even when AI systems correctly infer and follow human intent, they can still diminish human agency by making users increasingly predictable and dependent. They present the \emph{agency foundations} agenda as a framework for measuring and preserving human agency in AI systems. Similarly motivated, \cite{shen2024towards} proposes a framework for ``bidirectional human-AI alignment,'' emphasizing the necessity for mutual adaptation and alignment between AI systems and humans. Lastly, \cite{pedreschi2024human} explores how AI-driven recommendation systems shape human preferences, outlining some key challenges in measuring and mitigating these feedback loops.

While the dynamics between humans and AI are critical, we argue that describing the emergent behavior of an agentic AI system requires considering the dynamics at all interfaces (human-agent, agent-agent, and agent-environment).

\emph{AI design.} Regarding the design of AI systems themselves, \cite{huang2024position} discusses the integration of large foundation models with embodied systems through six core components: learning, memory, perception, planning, cognition, and action. They propose the \emph{agent foundation model}, incorporating multimodal reasoning and contextual memory to enhance prediction and adaptability. More broadly, \cite{johnson2024imagining} argues that existing AI systems lack ``wisdom": the ability to navigate intractable problems that involve radical uncertainty and ambiguity. They advocate for the development of metacognitive strategies (uncertainty estimation, self-reflection, and multi-perspective reasoning) to complement task-level problem-solving techniques. 

Regarding \cite{huang2024position}, we share the same philosophy: that agentic AI development would benefit from a more holistic view. Our position explores higher-level behavioral dynamics at the various interfaces of an agentic system as opposed to emergent properties at the lower architectural levels. Our position complements that of \cite{johnson2024imagining}, reinforcing that, if\footnote{Recall our earlier comment concerning our stance.} we wish to make agentic AI more capable, equipping it with more human qualities (like metacognition) can help. We argue, however, that such properties do not necessarily need to be embedded at the model level, rather they can emerge due to the (complex) interaction dynamics present in the system.

{\bf Outline.} The remainder of the paper is organized as follows. 

\emph{Defining Agentic Systems:} Section \ref{sec:def} presents our conceptualization of an agentic system. We introduce our working definition of agency, termed \emph{functional agency}, based on an existing decision-theoretic characterization. We argue that effective agentic systems are ones that possess a high degree of functional agency. 

\emph{Mechanisms of Emergence:} Section \ref{sec:emergence} describes some key mechanisms for emergence of system capabilities that exceed those of the system's individual components. By drawing on a significant amount of existing literature from various fields, we argue how interaction dynamics (both with the environment and among agents) can elevate the level of functional agency of the system as a whole. 

\emph{Open Challenges:} Informed by the mechanisms of emergence, Section \ref{sec:challenges} presents some key open challenges in building safe and effective agentic AI. 

\emph{Closing Remarks:} Lastly, in Section \ref{sec:alternative}, we provide some concluding remarks and offer general guidance for the design of agentic AI.

%\emph{Alternative Views \& Closing Remarks.} Lastly, in Section \ref{sec:alternative}, we situate our position by presenting an alternative view (and offer a counterpoint). Concluding remarks and general guidance for the community are provided.

%%%%%%%%%%%%%%%%%%%%%%%%%%%%%%%%%%%%%%%%%%
%%%%%%%%%%%%%%%%%%%%%%%%%%%%%%%%%%%%%%%%%%
%%%%%%%%%%%%%%%%%%%%%%%%%%%%%%%%%%%%%%%%%%
%%%%%%%%%%%%%%%%%%%%%%%%%%%%%%%%%%%%%%%%%%
%%%%%%%%%%%%%%%%%%%%%%%%%%%%%%%%%%%%%%%%%%
%%%%%%%%%%%%%%%%%%%%%%%%%%%%%%%%%%%%%%%%%%
%%%%%%%%%%%%%%%%%%%%%%%%%%%%%%%%%%%%%%%%%%
%%%%%%%%%%%%%%%%%%%%%%%%%%%%%%%%%%%%%%%%%%

\section{Defining Agentic Systems}
\label{sec:def}

Constructing any systems theory requires clarity of definitions, boundaries, and the nature of the interactions among the system's components. We first state our definition of \emph{agency} in the context of AI systems, then describe our conceptualization of an \emph{agentic system} in terms of the interaction between humans, agents, and the external environment.

\subsection{Agency}

Efforts to define agency date back to ancient philosophy, particularly the works of Aristotle, who explored the concept of causality and intentionality in his works \emph{Metaphysics} and \emph{Nicomachean Ethics}. In modern times, agency has been extensively studied (and debated) within the fields of psychology (e.g., self-efficacy \cite{bandura1982self,bandura2001social}), sociology (e.g., structuration theory \cite{giddens1984constitution}), philosophy (e.g., intentionality \cite{dennett1989intentional}), and biology (e.g., boundaries of ``self'' \cite{levin2019computational, fields2022competency}). The general consensus is that agency describes an entity's capacity to act independently, make decisions, and influence its environment in pursuit of goals or objectives, with differences mainly centered on the degree of intentionality, purposiveness, and autonomy in doing so.

The type of agency used in the AI community to discuss agents differs markedly from how agency is understood in discussions of human behavior and cognition. Generally, the conditions of AI agency discussed in the AI community are significantly looser than those applied to human agency. Early definitions describe an agent as a ``system that tries to fulfill a set of goals'' \cite{maes1993modeling}, or ``anything that can be viewed as perceiving its environment through sensors and acting upon that environment through effectors'' \cite{russell1995artificial}, to more comprehensive definitions as a ``system situated within and a part of an environment that senses that environment and acts on it, over time, in pursuit of its own agenda and so as to effect what it senses in the future'' \cite{franklin1996agent}. As argued by \cite{barandiaran2009defining}, many of these definitions are incomplete and ``rely on additional undefined terms'' like sensing, perception, action, and goal. More recent definitions in the context of agentic AI, have not helped much to resolve this incompleteness, proposing definitions that primarily add the conditions that an agent is able to decompose a complex task into actionable subtasks and execute it with limited human supervision \cite{shavit2023practices,chan2023harms,bansal2024challenges,wiesinger2025agents,mitchell2025agents}. Generally, there is significant ongoing discussion on the precise conditions for AI agency \cite{barandiaran2024transforming,rouleau2024discussions}.

We adopt a definition of agency based on a causal definition, grounded in decision theory, from \cite{kenton2023discovering}. While stated relatively informally as ``systems that would adapt their policy if they were aware that their decisions influenced the world in a different way'', the statement 
points to a functional (rather than phenomenal \cite{chalmers1997conscious}) characterization of agency while still sharing some important aspects with genuine (human) agency \cite{rosenblueth1943behavior,emirbayer1998agency}.

\begin{definition}[Functional agency]\label{def:agency}
A system possesses \emph{functional agency} if the following three conditions are satisfied: 
\begin{itemize}
\item[i)] \emph{Action generation}: capable of generating actions, based on information from the environment, in the direction of some objective. 
\item[ii)]  \emph{Outcome model}: capable of representing relationships between actions and outcomes.
\item[iii)] \emph{Adaptation}: capable of adapting behavior in response to changes in the outcome model in a way that maintains or improves performance toward the objective.
\end{itemize}
\end{definition}

\renewcommand{\arraystretch}{1.3}
\begin{table*}[t!]
    \centering
    \begin{tabular}{m{2.5cm}m{4.3cm}m{4.3cm}m{4.3cm}}
%    \toprule
        \multicolumn{1}{c}{} & \multicolumn{1}{c}{\centering action generation} & \multicolumn{1}{c}{\centering outcome model} & \multicolumn{1}{c}{\centering adaptation} \\[0.2em]
    \midrule
    thermostat & \sethlcolor{yellowgreen}\hl{reactive}: decisions to heat/cool based on temperature measurements & \sethlcolor{lightred}\hl{none}: model implicit in design; physics of temp. change encoded in environment & \sethlcolor{lightred}\hl{none}: fixed heating/cooling behavior based on temperature thresholds \\
        \rowcolor{gray!10} autonomous car & \sethlcolor{lightgreen}\hl{stateful}: steers/brakes based on vehicle and environment state (inferred from sensors)  & \sethlcolor{lightgreen}\hl{intervention}: models how steering/braking influences position and speed & \sethlcolor{yellowgreen}\hl{contextual}: adapts driving behavior based on environmental conditions \\
    robotic gripper & \sethlcolor{lightgreen}\hl{stateful}:  specifies grasp forces/movement based on estimated object position  & \sethlcolor{lightgreen}\hl{intervention}: models how grasp force/motion influences object movement & \sethlcolor{lightgreen}\hl{parametric}: updates grasp policy parameters based on success/failure feedback \\
    \rowcolor{gray!10} LLM & \sethlcolor{lightgreen}\hl{stateful}: generates responses based on context state maintained during session & \sethlcolor{yellowgreen}\hl{association}: possesses correlations between prompts and responses & \sethlcolor{yellowgreen}\hl{contextual}: uses context to adapt information processing via attention patterns \\
    human & \sethlcolor{lightblue}\hl{epistemic}: actions driven by flexible knowledge structures and beliefs & \sethlcolor{lightblue}\hl{counterfactual}: ability to imagine and reflect on hypothetical scenarios & \sethlcolor{lightblue}\hl{reflective}: ability to evaluate and modify learning based on context and past experience \\
    \bottomrule
    \end{tabular}
    \caption{Varying degrees of functional agency as dictated by hierarchies of action generation (reactive $\rightarrow$ stateful $\rightarrow$ epistemic), outcome modeling (association $\rightarrow$ intervention $\rightarrow$ counterfactual), and adaptation ability (contextual $\rightarrow$ parametric $\rightarrow$ reflective).}
    \label{tab:examples}
\end{table*}

Action generation requires the system to be able to specify actions (as a function of information from the environment) toward a given objective, e.g., via a policy \cite{bellman1957dynamic,sutton1998reinforcement}. 
The ability to generate actions in the direction of an objective is core to agency, as it differentiates goal-directed behavior from undirected or habitual behavior \cite{balleine1998goal,gollwitzer1999implementation,bandura2001social,davidson2001essays,dolan2013goals,xu2024towards}. 
Generating such actions requires a model for how actions relate to outcomes in the environment (via the outcome model). Since goal-directed behavior describes specifying actions that achieve specific outcomes, action generation fundamentally depends on such a model \cite{bratman1987intention,von2004explanation,schlosser2019agency,niu2023influence,da2024active,richens2024robust}. Lastly, adaptation requires that the system is able to modify its behavior when the relationship between actions and outcomes changes. Without adaptation, the system would be unable to maintain goal-directed behavior over time \cite{varela1984living,di2005autopoiesis,thompson2011living}. Functional agency describes a type of autonomy in \emph{means} with respect to a specified goal (e.g., solution autonomy), rather than the stronger condition of autonomy in \emph{ends} or generating one's goals (e.g., goal autonomy or \emph{normativity} \cite{barandiaran2009defining,barandiaran2024transforming}).

Functional agency is \emph{not} a binary notion but rather exists on a spectrum, as dictated by the sophistication of the action generation process, the outcome model, and the ability to adapt. 
Action generation, at its simplest level, is given by a memoryless or \emph{reactive} policy \cite{singh1994learning} that maps immediate observations from the environment to actions, e.g., a thermostat's heating/cooling actions are based on the current temperature relative to the desired setpoint. Beyond reactive policies, \emph{stateful} policies generate actions as a function of some fixed-domain summary or sufficient statistic of the system \cite{kumar1986stochastic,hauskrecht2000value,tavafoghi2018sufficient,tavafoghi2021unified}, e.g., steering/braking actions in an autonomous car as a function of the estimated vehicle state. At the highest level, actions are generated via an \emph{epistemic} process, driven by abstract, context-sensitive knowledge representations (not necessarily with a fixed domain \cite{spelke2007core,tenenbaum2011grow}), e.g., how humans maintain information and make decisions. 
The outcome model underlying the action generation process obeys Pearl's causal hierarchy \cite{pearl2009causality}, ranging in complexity from simple \emph{associations} (statistical correlations) to \emph{interventions} (effect of taking actions) to \emph{counterfactuals} (imagined scenarios if past actions had been different). For example, an LLM operates on correlations between prompts and responses whereas autonomous vehicles operate using interventional models of how actions influence states. 
Adaptation mechanisms range from \emph{contextual} (modifying behavior based on context, such as past interactions or inferred conditions), \emph{parameteric} (updating the functional relationship between states and actions), or \emph{reflective} (deeper reasoning/reflection on \emph{how} to update the functional relationship). LLMs possess contextual adaptation, adjusting responses based on conversation history without changing internal parameters. Advanced robotic grippers \cite{openai2018learning,xu2021adagrasp} exhibit parametric adaptation by adapting their policy to account for (and partially offset) changes in the outcome model (e.g., one of its grippers becoming less responsive to control inputs). Humans exhibit reflective adaptation, capable of changing strategies altogether, e.g., switching from trial-and-error to rule-based reasoning \cite{lieder2020resource}, or recognizing when a model is wrong and discarding it when a viable alternative is discovered \cite{kuhn1997structure}. 
% Humans represent the highest degree of functional agency, possessing epistemic decision-making, counterfactual reasoning, and the ability to metacognitively adapt to changes. 
Functional agency naturally excludes devices that cannot adapt to changes in the outcome model (e.g., a thermostat) and objects that achieve outcomes completely by accident. Table \ref{tab:examples} outlines the degree of functional agency for some example systems.

\begin{figure*}[t!]
\centering
        \includegraphics[width=\textwidth]{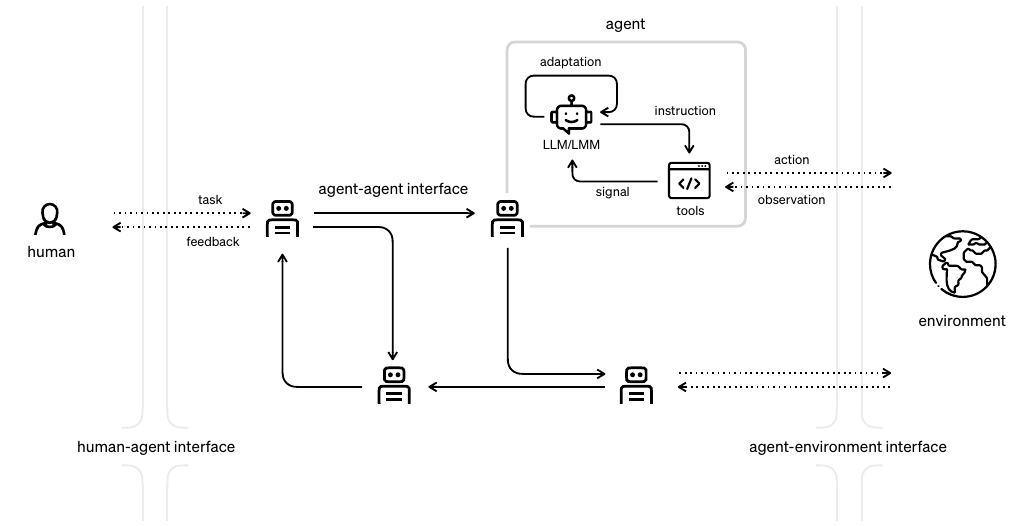}
    \caption{\emph{An agentic system}. The human user is responsible for seeding the initial task description and providing any feedback (in the form of clarification or approval) during the solution process. Each agent is described by an LLM or an LMM (large multimodal model), with access to tools that facilitate interaction with the external environment via actions (generated via instructions from the LLM/LMM) and observations (generating LLM/LMM-readable signals). These signals inform the agent's outcome model and drive any necessary adaptation. Agents additionally interact with other agents, communicating any relevant information about the task or observations from the environment.} 
    \label{fig:agentic_system}
\end{figure*}

\subsection{Agentic Systems}

An agentic AI system, or simply \emph{agentic system}, depicted in Fig. \ref{fig:agentic_system}, is a collection of agents interacting with humans and the environment with the objective of fulfilling specified goals. Practically, an agent is an LLM or large multimodal model (LMM) with access to tools --- specialized components/functionalities like APIs, external services, computational resources, or domain-specific software --- that allow it to perform specific operations in the environment.\footnote{Tools have no agency and must be explicitly invoked with well-defined parameters.} In this sense, tools define both the capabilities (actions) of the agent and the information (via observations/signals) that can be obtained from the environment. The human is responsible for seeding the initial task specification, providing clarification\footnote{The theories of (incomplete) contracts and bounded rationality imply a fundamental impossibility of specifying preferences across all possible contingencies a priori \cite{simon_1957,williamson1975markets,grossman1986costs}.}, and authorizing any (agent) actions that need human approval \cite{shavit2023practices}. Given the task specification, an agent is able to interact with other agents (agent-agent interaction) to facilitate task decomposition/planning and delegation. This interaction can be cooperative or competitive, e.g., in the case of limited compute. The environment consists of everything external to the agentic system. This includes infrastructure (computers), other humans, other agents, and even other agentic systems. In this sense, our treatment of the environment in an agentic system is similar to that in RL, where it encompasses all elements that can influence or be influenced by the agents' actions.

We claim that effective agentic systems, as measured by their ability to carry out complex tasks in novel settings, are those that exhibit a high degree of functional agency. Many of the limitations of modern LLMs (and consequently agents) can be described by two factors: i) their inability to causally reason \cite{zevcevic2023causal,jin2023can,romanou2023crab} and, ii) their lack of metacognitive awareness \cite{johnson2024imagining,griot2025large}. First, while current LLMs can effectively mimic causal behavior in familiar settings, they lack true causal reasoning \cite{zevcevic2023causal}, often facing difficulty distinguishing correlation from causation\footnote{This manifests as a form of reasoning brittleness in which their causal inference abilities are restricted to ``in-distribution settings when variable names and textual expressions used in the queries are similar to those in the training set'' \cite{jin2023can}.} and struggling with complex causal structures \cite{romanou2023crab}.\footnote{Various benchmarks validate this behavior \cite{kapkicc2024introducing,zhou2024causalbench,yang2024critical}.} Second, current LLMs lack metacognitive awareness \cite{johnson2024imagining}, such as misunderstanding their own goals in open-ended settings \cite{li2024think}, experiencing ``metacognitive myopia'' in evaluating source validity and handling of repetitive information \cite{scholten2024metacognitive}, and exhibiting systematic overconfidence, providing assured answers even when lacking sufficient information \cite{griot2025large}. These deficiencies can be characterized by a low degree of functional agency, namely from a lack of both epistemic monitoring --- the ability to detect inconsistencies and recognize when additional reasoning is required --- and control --- the ability to update beliefs and adapt behavior, via reflection, in response to detected errors \cite{nelson1996consciousness,thompson2011intuition,ackerman2017meta,scholten2024metacognitive}.

The essence of the systems view is that it is not necessary for every component to be highly functionally agentic for the system as a whole to possess a high level of functional agency. Tool use, the capacity to maintain a state (or local memory), and the ability to interact with the environment and other agents can lead to a collective agency beyond that of the individual components.

%%%%%%%%%%%%%%%%%%%%%%%%%%%%%%%%%%%%%%%%%%
%%%%%%%%%%%%%%%%%%%%%%%%%%%%%%%%%%%%%%%%%%
%%%%%%%%%%%%%%%%%%%%%%%%%%%%%%%%%%%%%%%%%%
%%%%%%%%%%%%%%%%%%%%%%%%%%%%%%%%%%%%%%%%%%
%%%%%%%%%%%%%%%%%%%%%%%%%%%%%%%%%%%%%%%%%%
%%%%%%%%%%%%%%%%%%%%%%%%%%%%%%%%%%%%%%%%%%
%%%%%%%%%%%%%%%%%%%%%%%%%%%%%%%%%%%%%%%%%%
%%%%%%%%%%%%%%%%%%%%%%%%%%%%%%%%%%%%%%%%%%

%%%%%%%%%%%%%%%%%%%%%%%%%%%%%%%%%%%%%%%%%%
%%%%%%%%%%%%%%%%%%%%%%%%%%%%%%%%%%%%%%%%%%
%%%%%%%%%%%%%%%%%%%%%%%%%%%%%%%%%%%%%%%%%%
%%%%%%%%%%%%%%%%%%%%%%%%%%%%%%%%%%%%%%%%%%
%%%%%%%%%%%%%%%%%%%%%%%%%%%%%%%%%%%%%%%%%%
%%%%%%%%%%%%%%%%%%%%%%%%%%%%%%%%%%%%%%%%%%
%%%%%%%%%%%%%%%%%%%%%%%%%%%%%%%%%%%%%%%%%%
%%%%%%%%%%%%%%%%%%%%%%%%%%%%%%%%%%%%%%%%%%

\section{Mechanisms of Emergence}
%\section{Mechanisms of Collective Agency}
%\section{Agentic Systems Theory}
\label{sec:emergence}

Emergence is driven by interactions at all scales of an agentic system. In what follows, we describe mechanisms of emergence for some fundamental capabilities.

\subsection{Environment enhances cognition} 
\label{ssec:cognition}

Perhaps the most direct mechanism of emergent capabilities is via \emph{embodied cognition} \cite{merleau1945phenomenology,varela1991embodied,barsalou1999perceptual} --- the principle that cognitive processes are shaped by interactions with the environment rather than being purely abstract mental computations. In the case of human development, enhanced cognition arises from sensorimotor activity, namely the coordinated interaction of multiple sensory and motor systems through physical exploration and manipulation of the environment \cite{ballard1997deictic,smith2005development}. In an agentic system, enhanced cognition arises due to the agent's interaction with the environment via tools, effectively acting as the ``sensorimotor'' interface that enables the agent to perceive and manipulate its environment. We outline some key mechanisms from developmental psychology for how interaction with one's environment can lead to emergent capabilities.

{\bf Generalized representations from multimodality.} One of the primary reasons that sensorimotor interaction with the environment aids cognition is due to \emph{multimodality} \cite{smith2005development}. When multiple modalities provide correlated information about the same phenomenon, the brain combines these signals (via \emph{reentrant neural maps} \cite{edelman1987neural}) to detect/correct errors and form abstract representations that capture invariant properties across modalities. 
For instance, the concept of ``roundness'' emerges from the correlation between visual curvature, tactile smoothness, and the motor patterns needed to trace a circular path. This allows for the discovery of ``higher-order regularities that transcend particular modalities'' and facilitates powerful learning capabilities \cite{smith2005development}. Importantly, such discovery can take place entirely via observation of one's own actions without the need for assigned tasks or teachers \cite{piaget1952origins,bushnell2013dual}.

In an agentic system, cross-referencing signals from multimodal signals would allow an agent to create stable representations of concepts in the environment, potentially aiding generalization ability. Multimodal models have already shown to aid learning \cite{huang2023embodied,li2025semgrasp}, demonstrating improved performance as a result of combining mutually reinforcing signals via methods like ``cross-modal transfer'' \cite{huang2023language}. As agents become more multimodal, as facilitated by multimodal models and associated tools  \cite{alayrac2022flamingo,sun2023generative,zhang2024mm}, we may begin to see agents with significantly enhanced cognition for the same reasons as in (multimodal) human cognition. The primary lesson is that designing an agent is not simply about deliberate design of its representational ability --- we must factor in the impact of the agent's (multimodal) interaction with its environment on its ability to form rich representations and learn.

{\bf Prematurity helps.} While the design of neural networks is inspired by biological processes, the way in which they learn (or are trained) differs fundamentally from how humans learn \cite{lake2017building}. Human babies are not given an enormous dataset of how the world works. Rather, their learning is exploratory and incremental \cite{gopnik1999scientist,gopnik2020childhood}, necessarily \emph{not} reliant on prior information. Their initial lack of sophistication, or \emph{prematurity}, is core to how they develop their cognitive abilities: regularities and correlations change as cognition develops, and capabilities emerge in a precise order \cite{smith2005development}.

A natural consideration for agentic systems is if a better path to generalized agents is to rely more on the abilities that agents develop through interaction with their environment, and less on the prior knowledge embedded via pretraining. In the context of RL agents, the main novelty of AlphaGo Zero \cite{silver2017mastering} over AlphaGo \cite{silver2016mastering} was its ability to learn entirely from self-play, in turn allowing for the emergence of more general strategies (via iterative self-improvement) without being influenced by the biases of existing human play/strategies. In the context of agentic AI, allowing a weakly pretrained agent to explore its environment (via multiple modalities) could be a viable path to generalized representations and abilities. Developmental robotics \cite{cangelosi2015developmental} and the study of intrinsic motivation (or curiosity-driven learning) \cite{barto2013intrinsic,oudeyer2007intrinsic,singh2010intrinsically}\footnote{See the intrinsic motivation and open-ended learning (IMOL) community \url{https://www.imol-community.org}.} may offer insights into deliberately designing such emergent abilities.

\subsection{Ability to predict enables reasoning}

The mechanisms for how causal reasoning emerges from simpler processes is an incredibly complex topic \cite{ellis2012top,gopnik2012reconstructing,hoel2013quantifying}. One compelling description from neuroscience describes the emergence of causal learning via the \emph{free-energy principle} \cite{friston2007free,friston2010free} and \emph{hierarchical predictive processing} \cite{clark2013whatever,hohwy2013predictive}. The core idea is that the brain constructs generative models for (top-down) predictions of sensory inputs and refines these predictions through (bottom-up) error signals from the environment. When prediction errors are observed, the system can either update its internal model, via \emph{perceptual inference} \cite{friston2010action}, or take actions to help make its predictions come true, via \emph{active inference} \cite{friston2003learning}. The main argument is that causal models emerge directly as a consequence of this progressive, error-minimizing refinement: the observation and (active) sampling of the environment creates a causal perception-action loop that identifies causal structures.

In an agentic system, an agent could hypothetically perform a similar error-minimization process to iteratively construct its causal model(s). Some tools have already enabled agents to actively sample their environment (via code execution \cite{hu2024automated}) and perform complex experiments \cite{narayanan2024aviary,huang2024crispr}, both necessary components of this process. Agentic systems are not currently known to employ explicit hierarchical predictive processing methods, however, the simplicity of the process (simply minimizing prediction errors) indicates that this mechanism could become a viable path to emergent causal reasoning in agentic systems.

\subsection{Prediction and interaction enables metacognition}

Effectively adapting to changes requires reasoning/reflection about the underlying process that led to that change. Such metacognitive reasoning emerges from similar mechanisms as that of causal reasoning and is amplified by interaction with others. Namely, error detection is argued to emerge directly from a model inferring that its action was incorrect (given available evidence), as measured by the disagreement between the decision variable and the confidence variable, and does not require an ``explicit error detection mechanism'' \cite{yeung2004neural, fleming2017self}. Social interaction (or collaboration) enhances this process by enabling individuals to calibrate their confidence estimates against estimates of the group \cite{bahrami2010optimally,bang2018distinct,surowiecki2004wisdom}. This yields \emph{shared representations} --- internal models that encode both individual and group-level confidence signals --- allowing for more effective coordination \cite{frith2012mechanisms,shea2014supra,wolf2023shared} and ultimately the ability for individuals to efficiently and intelligently adapt (in the direction of fewer errors) to changes in the environment \cite{wegner1987transactive,holland1992complex,hutchins1995cognition,simon2012architecture}.

In an agentic system, allowing agents to form predictions (with associated confidences) of concepts in their environment (e.g., via tools), and additionally facilitating communication of these uncertainties to other agents, can allow for the formation of such shared representations and the emergence of metacognitive awareness. Current efforts to incorporate uncertainty quantification into LLMs \cite{lin2023generating,balabanov2024uncertainty,shorinwa2024survey} and architectures that allow for agent-to-agent communication \cite{li2024survey}, provide a viable path for collective metacognitive behavior. Importantly, this behavior can arise without intentionally designing it at the individual agent level, but rather as a result of lower-level behaviors like (contextually) adapting to changes, quantifying and maintaining uncertainties (via local memory or state), and communicating these uncertainties to other agents.

%%%%%%%%%%%%%%%%%%%%%%%%%%%%%%%%%%%%%
%%%%%%%%%%%%%%%%%%%%%%%%%%%%%%%%%%%%%
%%%%%%%%%%%%%%%%%%%%%%%%%%%%%%%%%%%%%
%%%%%%%%%%%%%%%%%%%%%%%%%%%%%%%%%%%%%

\section{Open Challenges}
\label{sec:challenges}

The mechanisms discussed in Section \ref{sec:emergence} require further developments to realize and, importantly, give rise to various risks if implemented. In this section, we outline some key open challenges in the development of effective and safe agentic AI systems.

\subsection{Building generalist agents}

There are many practical questions underlying the mechanisms outlined in the previous section. Regarding the emergence of generalized representations/abilities via multimodal interaction, to what level of pretraining is required to enable agents to meaningfully explore their environment? Insights from the development of the generalist agent \emph{Gato} \cite{reed2022generalist} indicate that while extensive pretraining across diverse tasks can aid an agent’s generalization ability, it may not be necessary for all forms of learning. Experiments on (fine-tuning for) out-of-distribution tasks suggest that models can efficiently adapt with less pretraining provided they have structured mechanisms for exploration. In some cases (learning new Atari games) pretraining did not yield a clear advantage, implying that targeted exploration with the environment may be preferred over pretraining.

This raises several questions: Could an agent with minimal pretraining, but equipped with mechanisms for self-directed exploration and curiosity-driven learning, achieve superior generalization? If so, what forms of exploration (such as intrinsic motivation, goal-directed play, or unsupervised environment modeling) would be most effective? How does the balance between pretraining and learning depend on the complexity (and diversity) of the task distribution? In the event that an agent is able to learn skills in-situ, in what order does the agent develop capabilities? Can this order be influenced to improve generalization ability?

\subsection{Designing efficient agent-agent interactions}

A core feature of agentic systems is their ability to decompose complex tasks into subtasks and delegate them among the agents \cite{zhu2024redel}. Doing so efficiently requires understanding both the dependencies between the subtasks (the order in which they need to be completed) and which agents are most capable at which subtasks. In human systems, tasks are decomposed and delegated to others based on inferred capabilities given evidence from previous experience. Humans often maintain trust not only on specific tasks but on general categories of tasks (e.g., successfully writing Python code on one project likely implies ability to write effective Python code on an unrelated project).\footnote{This is an instance of the halo effect bias \cite{thorndike1920}.}

A key question in the delegation of subtasks in an agentic system is to what degree should trust on a given task transfer to trust on a different task? The precise trade-off is unclear: overly relying on evidence from specific tasks would lead to significant data sparsity and inability to delegate, whereas transferring trust too generously would lead to suboptimal task outcomes. What features of tasks and agents influence the appropriate trade-off? How should the cold-start problem (delegation of a new task or to a new agent) be addressed? Structures from organizational management \cite{lai2017systems,denning2022digital}, e.g., hierarchy of authority versus network of competence, may inform general strategies for how to decompose/delegate diverse tasks among agents.

\subsection{Controlling emergence of subgoals}  % risks

The ability of agents to decompose and delegate subtasks to other agents can lead to emergence of a higher degree of autonomy. For example, even in a simple system with two agents, one agent can decompose the original task and assign a subtask (or subgoal) for the other agent. Collectively, this two-agent system possesses a degree of goal autonomy beyond the solution autonomy of each, simply due to the first agent defining the goal for the second agent. The more complex the initial (human-seeded) task, and the more agents that exist in the system, the longer the chain of potential subgoals. While the initial task partially constrains the overall task outcome, the constraint imposed by the human's initial task specification on intermediate subgoals becomes weaker as the chain grows in length.

One fundamental challenge is how the generation of these subgoals should be monitored. The intended speed and scale at which agentic systems will be deployed precludes full reliance on humans for the monitoring. However, relying on another agent to monitor subgoal creation brings us back to the original problem. What monitoring structures are most effective? What role can humans play? Does limiting an agentic system's ability to create subgoals reduce its ability to successfully carry out tasks? If so, how should agents be incentivized to not evade this monitoring?

\subsection{Governing human-agent interactions}

A contributing factor in the emergence of unsafe subgoals is the user's inability to specify what is ``safe'' across all possible contexts and contingencies. This is an unavoidable property of communication and arises due to fundamental bounds on rationality \cite{simon_1957,williamson1975markets}. In traditional settings, namely incomplete contracts \cite{grossman1986costs,hart1990property}, underspecification is addressed through \emph{residual control rights}, which determine who has decision-making authority in situations not explicitly covered by the contract \cite{hart1990property,hart1995firms}. Determination of these rights is typically dictated by the parties' relative bargaining power, risk allocation, available information, and expertise \cite{aghion1992incomplete,aghion1997formal,baker2002relational}.

The design of residual control rights for agentic systems may be an effective strategy for mitigating risks. Fundamental differences in capabilities between humans and agents point to some natural divisions in control rights. Agents should retain control over highly time-constrained local decisions (e.g., evasive maneuvers), computationally-intensive tasks, well-defined routine decisions with clear metrics (and bounded risk), and decisions that rely on information only available at the agent-environment interface. One issue is that a sequence of many low risk, but automated, agent decisions may create larger emergent risks over time. How can the accumulation of risk from sequences of local decisions be reliably detected? Humans should retain control over longer-term strategic decisions, novel tasks requiring value judgments, and decisions with significant (or irreversible) safety risks. In the event that the agent is uncertainty about a decision, escalation mechanisms could be designed that handoff the decision to a human. How can decisions be escalated to a human in order to allow enough time to interpret the available information and take an action? Research on human-agent communication may provide useful insights \cite{bansal2024challenges,burton2024large}.

\section{Closing Remarks}
\label{sec:alternative}

We have argued that the development of agentic AI is in need of a systems view in order to accurately estimate both capabilities and risks. Our position is grounded in a definition of agency, termed functional agency, that quantifies the degree of agency of a system by its ability to take goal-directed actions, model outcomes, and adapt behavior (in the direction of the goal) when the action-outcome model changes.\footnote{This definition of agency contributes to the growing body of literature on AI agency \cite{kenton2023discovering,barandiaran2024transforming,abel2025agency,zhang2025conceptualizing}.} We argue that effective agentic systems are those that possess a high level of functional agency.  

The primary philosophy of the systems view is that a system can possess a high level of functional agency simply due to the (complex) interactions in the system, notably even when individual agents are much simpler. Informed by a large amount of literature from various fields (psychology, neuroscience, cognitive science, sociology, and biology), we outline some viable pathways in which functional agency can emerge: i) enhanced cognition due to an agent's interaction with its environment, ii) emergence of causal reasoning due to the ability to minimize prediction errors, and iii) emergence of metacognitive awareness due to the ability to predict, quantify uncertainty, and communicate with other agents. These mechanisms hint at possible emergent capabilities in agentic systems.

While we argue that there are viable paths for emergent capabilities, we are not saying these are automatic; we must design/facilitate the properties that these mechanisms rely on. We must understand the mechanisms of emergence in order to intentionally design such properties into agentic systems and to limit the associated risk. Additionally, to reiterate a previous point, we are not advocating for the unconstrained development of increasingly agentic systems.\footnote{See \cite{mitchell2025fully} for a similar view.} Rather, we argue that understanding these emergent capabilities provides the AI community with essential tools for mitigating their risk.\footnote{See \cite{kolt2025lessons} for discussion of some prominent risks.} We believe that the systems-level view can lead to identification of many more mechanisms not discussed in our paper.

These considerations will become increasingly important as advancements in AI continue to progress. The discussion in our present paper was largely restricted to current-day LLMs/agents that interact with the world via text. We as a community need to consciously consider the impact of additional modalities, e.g., speech, vision, touch/movement (via a robotic ``body''), on the overall cognitive abilities of the system. Such considerations will help to ensure that AI safely and effectively augments human capabilities while preserving human agency.

%%%%%%%%%%%%%%%%%%%%%%%%%%%%%%%%%%%%%%%
%%%%%%%%%%%%%%%%%%%%%%%%%%%%%%%%%%%%%%%
%%%%%%%%%%%%%%%%%%%%%%%%%%%%%%%%%%%%%%%
%%%%%%%%%%%%%%%%%%%%%%%%%%%%%%%%%%%%%%%

\bibliography{references}
\bibliographystyle{icml2025}

\end{document}